%% file: aaai2026.tex
\documentclass[letterpaper]{article} % DO NOT CHANGE THIS
\usepackage{aaai2026}  % DO NOT CHANGE THIS
\usepackage{times}  % DO NOT CHANGE THIS
\usepackage{helvet}  % DO NOT CHANGE THIS
\usepackage{courier}  % DO NOT CHANGE THIS
\usepackage[hyphens]{url}  % DO NOT CHANGE THIS
\usepackage{graphicx} % DO NOT CHANGE THIS
\usepackage{color}
\usepackage{natbib}  % DO NOT CHANGE THIS AND DO NOT ADD ANY OPTIONS TO IT
\usepackage{caption} % DO NOT CHANGE THIS AND DO NOT ADD ANY OPTIONS TO IT
\usepackage{algorithm}
\usepackage[noend]{algorithmic}
\usepackage{newfloat}
\usepackage{listings}
\DeclareCaptionStyle{ruled}{labelfont=normalfont,labelsep=colon,strut=off} % DO NOT CHANGE THIS
\floatstyle{ruled}
\newfloat{listing}{tb}{lst}{}
\floatname{listing}{Listing}
\usepackage{amsmath}
\usepackage{amssymb}
\usepackage{amsthm}
\usepackage{cleveref}
\usepackage{subcaption}
\usepackage{comment}
\usepackage{multirow}
\usepackage{makecell}
\usepackage{array}
\usepackage{mathabx}
\usepackage{makecell}
\usepackage{booktabs}

\DeclareMathOperator*{\Argmax}{arg\,max}

\DeclareMathOperator{\clip}{clip}
\newcommand{\clipped}[1]{\underline{\bar #1}}
\newcommand{\Clipped}[1]{\underline{\overline #1}}

\title{SMiLE: Provably Enforcing Global Relational Properties in Neural Networks}
\author {
    Matteo Francobaldi\textsuperscript{\rm 1},
    Michele Lombardi\textsuperscript{\rm 1},
    Andrea Lodi\textsuperscript{\rm 2}
}
\affiliations {
    \textsuperscript{\rm 1}DISI, University of Bologna \\
    \textsuperscript{\rm 2}Jacobs Technion-Cornell Institute, Cornell Tech and Technion\\
    \{matteo.francobaldi2, michele.lombardi2\}@unibo.it, andrea.lodi@cornell.edu
}
\begin{document}

\maketitle 

%%%%%%%%%%%%%%%%%%%%%%%%%%%%%%%%%%%%%%%%%%%%%%%%%%%%%%%%%
\begin{abstract}
% Artificial Intelligence systems, particularly Neural Networks (NNs), are increasingly adopted in high-stakes domains, where ensuring the satisfaction of critical properties, such as robustness, fairness, or domain-specific constraints, is essential for compliance with regulatory frameworks and alignment with human values. 
% However, enforcing properties in NNs is very challenging, due to the non-linear and high-dimensional nature of these models, this is why existing methods are either limited to specific (mostly local) properties, or fail to guarantee their validity. 
% In this work, we tackle these limitations by extending SMiLE, a recently proposed enforcement framework for NNs, to support \emph{global relational properties}, broadening its scope and applicability. 
% The proposed approach scales well with model complexity, accommodates general properties and architectures, and provides full property satisfaction guarantees. 
% We evaluate SMiLE on \emph{monotonicity}, \emph{global robustness} and \emph{individual fairness}, enforced both on synthetic and real-world data, for both regression and classification tasks. 
% Our approach is competitive with -- or even superior to -- property-specific baselines, in terms of both accuracy and runtime.
% Overall, our contributions establish a general-purpose, scalable framework for provably enforcing global properties in neural networks, opening up promising future research and applications.
Artificial Intelligence systems are increasingly deployed in settings where ensuring robustness, fairness, or domain-specific properties is essential for regulation compliance and alignment with human values. 
However, especially on Neural Networks, property enforcement is very challenging, and existing methods are limited to specific constraints or local properties (defined around datapoints), or fail to provide full guarantees. 
We tackle these limitations by extending SMiLE, a recently proposed enforcement framework for NNs, to support global relational properties (defined over the entire input space). The proposed approach scales well with model complexity, accommodates general properties and backbones, and provides full satisfaction guarantees. 
We evaluate SMiLE on monotonicity, global robustness, and individual fairness, on synthetic and real data, for regression and classification tasks. 
Our approach is competitive with property-specific baselines in terms of accuracy and runtime, and strictly superior in terms of generality and level of guarantees. 
Overall, our results emphasize the potential of the SMiLE framework as a platform for future research and applications.
\end{abstract}

\begin{links}
    \link{Code}{https://github.com/Francobaldi/SMiLEAAAI2026}
    %\link{Extended Version}{}
\end{links}

\section{Introduction}
\label{sec:introduction}
\input{introduction}

\section{Related Work}
\label{sec:related_work}
\input{related_work}

%\section{Relational Properties}
%\label{sec:relational_properties}
%\input{relational_properties}

\section{Methodology}
\label{sec:methodology}
\input{methodology}

\section{Experimentation}
\label{sec:experimentation}
\input{experimentation}

\section{Conclusion}
\label{sec:conclusion}
\input{conclusion}

\section*{Acknowledgements}
The project leading to this application has received funding from the European Union’s Horizon Europe research and innovation programme under grant agreement No. 101070149.
 
\bibliography{aaai2026}

\clearpage

\clearpage

\section*{Reproducibility Checklist}
\label{sec:reproducibility_checklist}
\input{reproducibility_checklist}

\input{supplemental}

\end{document}

%% file: introduction.tex
Artificial Intelligence (AI) has witnessed tremendous success in recent years, becoming a pervasive technology. 
This has been fueled in many domains by Machine Learning (ML) models such as Neural Networks (NN), and by purely data-driven algorithms.
While highly effective in terms of accuracy, these methods have difficulties in providing satisfaction guarantees on additional properties, which is critical when regulatory compliance or reliance on expert knowledge are necessary.
% however, the development of AI systems, particularly Machine Learning (ML) models such as Neural Networks (NNs), still relies on purely data-driven algorithms that solely optimize prediction quality, while overlooking properties that, depending on the use case, might be either strictly required or highly desirable.
% Broadly speaking, two main reasons motivate the need for enforcing properties in these systems: regulatory compliance and knowledge injection. 
The former involves safety-critical or ethically sensitive scenarios, where legal frameworks impose specific requirements, such as robustness in autonomous driving (e.g., resisting adversarial attacks in traffic sign recognition) or fairness in automated hiring (e.g., avoiding gender bias in candidate selection).
The latter applies to settings that, even though not necessarily critical, might still benefit from the incorporation of properties, like monotonicity in remaining useful life estimation (e.g., the condition of a machine can only deteriorate over time), or adherence to physical laws in scientific modeling (e.g., conserving mass or energy in fluid simulations).
% , or combinatorial constraints in the integration with reasoning frameworks (e.g., resource constraints in planning and scheduling).

% \paragraph{Problem Setting}

In this work, we consider properties expressible as universally quantified implications between two predicates $Q$ and $R$, specified over the input and output variables of a ML model $f\colon \mathcal{X} \rightarrow \mathcal{Y}$, that is: 
\begin{equation}
\label{eq:property}
\forall x_1,..,x_k \in \mathcal{X}, Q(x_1,..,x_k) \Rightarrow  R(f(x_1),..,f(x_k))
\end{equation}
where each $x_i$ represents a distinct input vector.
We distinguish properties according to two criteria, \emph{arity} and \emph{scope}. The arity is \emph{trace} if $k=1$, \emph{relational} otherwise. The scope is \emph{local} if $Q$ and $R$ depend on data-based information, \emph{global} if they apply also out-of-distribution. Our definition generalizes a broad range of requirements,
such as adversarial (local) robustness ($\|x - x_0\| \leq \delta \Rightarrow f(x)=f(x_0)$, with $x_0 \in D_{\text{train}}$), safety ($x\in S_{\text{in}} \Rightarrow f(x) \in S_{\text{out}}$), and monotonicity ($x' \leq x'' \Rightarrow f(x') \leq f(x'')$).
Global relational properties generally allow to express more advanced requirements, but they are also more difficult to handle, as they require searching over the entire domain, as well as reasoning over multiple model evaluations \cite{Banerjeinput2024,blatter2022certified}.
Finally, a property can be enforced with or without \emph{guarantees} on its validity, that is, on the non-existence of a counterexample where such property is violated. Guarantees are crucial in regulatory scenarios, while not always necessary in knowledge injection ones.

\paragraph{Enforcing Properties in ML} Regardless of the arity and scope, enforcing properties in complex AI systems such as NNs is challenging due to the non-linear and high-dimensional nature of these models \cite{katz2017reluplex}. 
The effort in this direction has been mostly spent on trace properties, in particular on adversarial robustness and safety, as surveyed by \cite{liu2021algorithm,meng2022adversarial,muhammad2022survey}. 
Recent years, however, have witnessed a shift of interest to relational ones, in particular fairness, robustness and monotonicity. 
Most of the proposed approaches embed carefully designed verifiers into training procedures, to promote property feasibility \cite{tumlin2024fairnnv,athavale2024verifying,benussi2022individual,liu2020certified,khedr2023certifair}. 
Others modify the model architecture to make it systematically feasible \cite{kitouni2023expressive,nguyen2023mononet,runje2023constrained,leino2021globally}. 

Despite the substantial progress achieved in the last decade, critical challenges remain, most notably the lack of guarantees and generality.
Most of the existing methods improve the degree of property satisfaction, but without certifying its validity.
Moreover, many of them are narrowly tailored to specific properties (e.g., adversarial robustness) or models (e.g., ReLU NNs).
To the best of our knowledge, the only existing frameworks capable of providing guarantees while retaining generality are \cite{goyal2024deepsade} and \cite{francobaldi2025smle}.
The two approaches, both limited to trace properties, are based on similar principles: during training they adjust the model weights to guarantee feasibility, in the former case by solving a Quantified SMT problem, in the latter by pairing counterexample generation and Projected Gradient Descent.
% that explicitly aims to address these issues (expanding the generality of the enforcement while providing full guarantees) is the one introduced by \cite{francobaldi2025smle}.

\paragraph{Contribution}

In this work, we extend the framework by \cite{francobaldi2025smle}, which was chosen due to the favorable trade-offs provided by its proposed architecture.
% This choice was motivated by the general neural architecture called SMiLE (Safe ML via Embedded overapproximation), 
This is referred to as SMiLE (Safe ML via Embedded overapproximation) and is obtained by augmenting an arbitrarily complex backbone network with a much simpler, trainable overapproximator: the former is used for inference, to enhance expressivity, the latter for enforcement, to improve scalability.
% via a projected gradient method, so to make the property-aware training tractable. 
% The method guarantees property satisfaction upon convergence. 
% In its current version, SMiLE only supports trace properties. 
Our main contribution is introducing support for relational properties in this framework, through an extensive re-elaboration of its key components.
% In this work, we extend this framework to also handle relational ones, through a substantial re-elaboration of its key components. 
We evaluate the approach on 3 use cases: 
\emph{monotonicity} on synthetic data, \emph{robustness} on MNIST and \emph{fairness} on Compas, Crime and Law.
% , three commonly adopted datasets in the fairness literature. 
Across these benchmarks, our method can consistently achieve property guarantees, while remaining competitive with (and sometimes outperforming) property-specific baselines in terms of accuracy and runtime.
Finally, we discuss several research directions opened up by our contributions.

%% file: related_work.tex
Two trends mainly arise from the AI literature accounting for properties: verification and enforcing methods.
The former aim to formally certify the validity of properties in already-trained models, mostly by relying on optimization and searching techniques -- such as Mixed-Integer Programming (MIP) or Satisfability Modulo Theory (SMT) -- to seek counterexamples that falsify the assertion. 
The latter attempt to directly incorporate the desired property into the model’s behavior, either by designing property-aligned architectures, or by using existing verifiers into counterexample-guided training loops, or into real-time defense mechanisms.

Most of the effort has been dedicated, thus far, to trace properties. 
Adversarial training methods, proposed for adversarial robustness  \cite{madry2018towards, zhang2019you, shafahi2019adversarial, wang2020Improving, zhang2020attacks, wong2020Fast, kim2021understanding,ijcai2021p591}, local fairness \cite{mohammadi2023feta,benussi2022individual}, and local  monotonicity \cite{liu2020certified}, work by training the model over a combination of the original datapoints and their worst counterexamples.
Other methods work by detecting and purifying, or rejecting, malicious inputs \cite{dhillon2018stochastic, samangouei2018defensegan, yang2019menet, pang2018towards, metzen2017detecting, xu2017feature}, or by correcting the output to enforce constraint satisfaction \cite{WABERSICH2021109597,
yu2022towards,mohammadi2023feta}.

While the literature on trace properties is extensive, approaches capable of handling global relational properties have started to recently emerge. 
From a verification perspective, these methods rely on the same core idea: reducing the relational predicate to a trace one, by encoding multiple independent copies of the same model (product network) within a single verification instance \cite{Banerjeinput2024,tumlin2024fairnnv,athavale2024verifying}. 
The resulting verifiers can then be employed for property enforcement during training, as in \cite{benussi2022individual} and \cite{khedr2023certifair}, which design MIP-based verifiers to promote fairness, and in \cite{liu2020certified}, which proposes a MIP-based verifier to guarantee monotonicity.
Another class of approaches modify the design of the network to make it systematically feasible. \cite{kitouni2023expressive,nguyen2023mononet,runje2023constrained} constrain the network weights or activations to provide monotonicity guarantees, while \cite{leino2021globally} discourages robustness violations by introducing a violation class.
Finally, our work also aligns with Neuro-symbolic AI, which integrates logical reasoning into NNs \cite{giunchiglia2022deep}.

Many of these approaches are designed for specific properties, rely on strong architectural assumptions, e.g., small ReLU networks, or fail to provide satisfaction guarantees.
To the best of our knowledge, only \cite{goyal2024deepsade} and \cite{francobaldi2025smle} offer generality and guarantees, but only for trace properties, while no existing method can handle general relational ones. 
This work addresses these limitations by proposing a method to enforce generic relational properties in arbitrary networks, providing full satisfaction guarantees while remaining competitive with property-specific baselines.

%% file: methodology.tex
\paragraph{Relational Properties}

Although theoretically applicable to the general setting defined in \Cref{eq:property}, in this work we formulate our method for a real-valued neural network $f \colon \mathcal{X} \rightarrow \mathbb{R}$, where $\mathcal{X} = \bigtimes_{i=1}^n [l_i, u_i] \subseteq \mathbb{R}^n$ with $l_i \leq u_i \; \forall i$, and a global 2-arity property of the form
\begin{subequations}
\label{eq:explicit_property}
\begin{align}
Q(x',x'') \equiv \;\; & \underline{\delta}_i \leq x'_i - x''_i \leq \overline{\delta}_i, \forall i, \\ R(f(x'), f(x'')) \equiv \;\; & \underline{\epsilon} \leq f(x') - f(x'') \leq \overline{\epsilon}, 
\end{align}
\end{subequations}
for given input and output bounds $\underline{\delta}, \overline{\delta}\in \mathbb{R}^m$ and $\underline{\epsilon},\overline{\epsilon} \in \mathbb{R}$.  
The most common relational properties targeted in the literature can be derived from  \Cref{eq:explicit_property} by configuring its parameters, as shown in \Cref{tab:properties}, where $\delta, \epsilon \geq 0$, $M \in \mathbb{R}^+$ is a sufficiently large value used to relax unnecessary bounds, while $\mathcal{P}, \mathcal{P}^c \subseteq [m]$ denote the set of sensitive features (protected for fairness and monotonic for monotonicity) and its complement, respectively.
Intuitively:
1) in the robustness case, bounded input variations should translate into bounded output variations;
2) in the fairness case, when only the protected features change, the corresponding output change should be limited;
3) in the monotonicity case, when the sensitive features increase, the output cannot decrease.

\input{tables/properties}

\paragraph{SMiLE} SMiLE is an enforcement framework consisting of a verification-friendly neural architecture and a dedicated training algorithm \cite{francobaldi2025smle}. 
A SMiLE architecture can be built by decomposing a network $f$ into an arbitrary embedding function $h$ and a linear output function $g$, then by embedding a trainable overapproximator in between them, consisting of \emph{lower and upper auxiliary models} $\underline{h}, \overline{h}$, and a \emph{clipping operator} $\clip(z; l; u) = \max(l, \min(u, z))$:
\begin{align}
    \label{eq:smle}
    g(\clipped{z}; \theta_g)
    \circ \clip(z; \underline{h}(x; \theta_{\underline{h}}); \overline{h}(x; \theta_{\overline{h}}))
    \circ h(x; \theta_h).
\end{align}
The model structure, depicted in \Cref{fig:smle_arch}, ensures that the input to the $g$ function is contained into the bounding box
\begin{equation}
    \Clipped{H}(x; \theta_{\underline{h}}, \theta_{\overline{h}}) = [\underline{h}_1(x), \overline{h}_1(x)] \times \ldots \times [\underline{h}_n(x), \overline{h}_n(x)],
\end{equation}
where $n$ is the size of the embedding vector.
\input{figures/smile}
Many neural architectures employed in AI research and real-world applications can be easily adapted to this structure, including classifiers, by focusing on their logit output.

The original SMiLE training algorithm relies on two main components: 1) a generator, which searches for a counterexample that violates the property, and 2) a projector, which minimally adjusts the model parameters to restore the counterexample feasibility.
Training proceeds in two phases.
In the first phase (simply referred to as training), the generator and projector routines are embedded into a standard gradient-based algorithm, and executed after each gradient step to maximize accuracy while maintaining feasibility.
In the second phase (referred to as post-training), the update step is deactivated, and only the generation-projection loop is repeated to eliminate residual violations. 
If this process is successful, the resulting SMiLE model provably satisfies the desired property without requiring further intervention.

The key idea of the framework is to exploit the SMiLE architecture to speed up the generation and projection phase, which would be otherwise intractable for standard networks \cite{katz2017reluplex}.
When executing these steps, the method relaxes the arbitrarily complex constraints $z = h(x;\theta_{h})$ into the simpler bounding box constraints $\underline{h}(x;\theta_{\underline{h}}) \leq \clipped{z} \leq \overline{h}(x;\theta_{\overline{h}})$, while preserving the output ones $f(x, \theta) = g(\clipped{z};\theta_{g})$. 
In other words, the complex $h$ is ignored and only the much simpler $\underline{h}, \overline{h}$ and $g$ are used.
Since $g$ is simple by construction, and $\underline{h}$ and $\overline{h}$ can be made simple by design, the complexity is \emph{freely controllable}.
Note that ignoring the exact component $h$ in favor of the overapproximation computed by $\underline{h}$ and $\overline{h}$ preserves verification soundness (and hence feasibility), but may result in over-enforcement.
% This drawback, however, is not critical when the focus is on enforcing properties at training time, as soundness alone is sufficient to provide feasibility guarantees.

\input{algos/training}

\paragraph{Reengineered Training}

The framework from \cite{francobaldi2025smle} is limited to trace properties.
While the main advantage of the SMiLE architecture (i.e., simplified verification) readily applies to relational properties, attempting to reuse its original training algorithm -- with straightforward adaptations -- proved entirely inadequate to the new setting from \Cref{eq:explicit_property}.
This motivated us to design a thoroughly reengineered training approach that is more general, scalable, and stable than its predecessor.
% The resulting procedure is still based on the idea of identifying and resolving counter examples.

The procedure is described in \Cref{alg:training} and it is articulated in three sections, referred to as \emph{pretraining}, \emph{training}, and \emph{posttraining}.
In the pretraining phase, the architecture is trained via gradient descent (\textsc{PrimalStep}) to approximately satisfy certain regularization properties that greatly reduce the chance of getting stuck in poor local optima.
The training phase relies on Dual Ascent to reduce both estimation errors and property violations.
This is done by iteratively:
1) generating pairs of examples, i.e., a counterexample, with large violation (\textsc{Generator}); if a counterexample from past iterations is available, we check its validity (\textsc{Resolved}) before generating a new one.
%\textcolor{red}{Not sure of the second part of the sentence, that starting with ``if no such". Does not seem connected clearly.}
2) Using the counterexample to define a Lagrangian penalty term that is incorporated in the loss function (\textsc{TrainLoss}).
3) Performing a regular gradient descent step, followed by a gradient ascent step over the penalty term multiplier (\textsc{DualStep}).
The dual step increases the contribution of the property violation in the loss function, so that it is prioritized by the gradient descent process, with the aim to reach 0-violation.
% Once 0-violation is reached, the current counterexample is resolved.
When this is not possible, in the posttraining phase we attempt to adjust the model weights so that no counterexample exists anymore. 
This process is also based on dual ascent, but implements a projection operator that does not depend on the training data.
% An example of the outcome of each phase of the process is depicted in \Cref{fig:synthetic_example}.
In what follows, we provide details on the design of the critical subroutines in the algorithm.
The remaining subroutines are described in the paper appendix.

\input{figures/synthetic_example}

%and its differences w.r.t. the original method.

% In this work, we extend the framework to relational properties, in particular to those defined in \cref{eq:explicit_property}. 
% Specifically, we retain the original architecture but substantially revise the training procedure.
% In what follows, we re-elaborate each SMiLE component to accommodate the increased property complexity, after briefly describing its original version.
% For a rigorous and complete description of the original framework, we refer to \cite{francobaldi2025smle}.
% The SMiLE training framework consists of 3 phases, detailed below and integrated in \cref{alg:training}.

\input{algos/pretrainloss}

\paragraph{Pretraining Loss}

When adapting SMiLE for relational properties, we observed that traditional weight initialization strategies had a tendency to cause $\underline{h}$ and $\overline{h}$ to flip (the lower bound becomes higher than the upper bound).
When this happens, the semantic of the clipping operator prevents most gradient components from being backpropagated, causing catastrophic training failure.

We address this issue by pretraining the model with a custom loss, outlined in \Cref{alg:pretrainloss}, and featuring two types of terms.
On the one hand, the terms summed in $L_{\text{acc}}$, specified via a traditional loss function $L_s$ (i.e., MSE or BCE), maximize the accuracy of \emph{all} the possible output pathways in the SMiLE computation graph, i.e., $g \circ h$, $g \circ \underline{h}$, $g \circ \overline{h}$.
On the other hand, $L_{\text{box}}$ penalizes overapproximation degeneracy, occurring when $\underline{h}, \overline{h}$ and $h$ violate the constraints $\underline{h}(x;\theta_{\underline{h}}) \leq h(x;\theta_h) \leq \overline{h}(x;\theta_{\overline{h}})$.
\Cref{fig:synthetic_example} shows the outcome of the pretraining phase for data generated according to a simple univariate function: it can be seen that all output pathways provide similar results, and no significant flipping occurs.
The original SMiLE algorithm included only a basic pretraining step, where the entire architecture was optimized for accuracy, which regularly led to training failure in our preliminary experiments.

\paragraph{Generator}
Intuitively, the counterexample generator needs to search for \emph{a pair} of input vectors violating the target property.
In practice, we wish to use the SMiLE overapproximator to disregard the backbone network $h$ and accelerate this process.
Hence, we actually search for a pair of compound \emph{input and embedding vectors} $(x',x'' \clipped{z}', \clipped{z}'') \in  \mathcal{X}^2 \times \Clipped{H}(x'; \theta_{\underline{h}}, \theta_{\overline{h}}) \times \Clipped{H}(x''; \theta_{\underline{h}}, \theta_{\overline{h}})$.
In other words, the embedding $\clipped{z}', \clipped{z}''$ are only required to be in the overapproximation boxes associated with $x'$ and $x''$.
This is done by solving the following problem:
\begin{subequations}
\label{eq:generator}
\begin{align}
    & \Argmax_{x', x'', \overline{\underline{z}}', \overline{\underline{z}}''} \;\gamma \\
    \text{s.t. } 
    & l \leq x', x'' \leq u \\ 
    & \underline{\delta} \leq x' - x'' \leq \overline{\delta} \\
    & \underline{h}(x'; \theta_{\underline{h}}) \leq \clipped{z}' \leq \max(\underline{h}(x'; \theta_{\underline{h}}), \overline{h}(x'; \theta_{\bar{h}})) \label{eq:generator_flipping1}\\
    & \underline{h}(x''; \theta_{\underline{h}}) \leq \clipped{z}'' \leq \max(\underline{h}(x''; \theta_{\underline{h}}), \overline{h}(x''; \theta_{\bar{h}})) \label{eq:generator_flipping2}\\
    & y' = g(\overline{\underline{z}}';\theta_g) \\
    & y'' = g(\overline{\underline{z}}'';\theta_g) \\
    & \gamma \leq y' - y'' -  \overline{\epsilon} + Mb \\
    & \gamma \leq -y' + y'' + \underline{\epsilon} + M(1-b) \\
    & x', x'' \in \mathbb{R}^{B^c} \times \{0,1\}^B, \clipped{z}', \clipped{z}'' \in \mathbb{R}^n, \; y', y'' \in \mathbb{R} \\
    & b \in \{0,1\}, \; \gamma \in \mathbb{R}_{>0}
\end{align}
\end{subequations}
where $M \geq 0$ is a fixed big-$M$ value, and $B,B^c \subseteq [m]$ respectively denote the set of binary variables and its complement, necessary when the learning task involves binary or one-hot encoded features.

Solving Problem \eqref{eq:generator} is dramatically easier than searching for counterexamples considering the actual backbone $h$, but can still be challenging, especially if moderately complex auxiliary models $\underline{h}, \overline{h}$ are used or if the model input is high-dimensional.
We propose to speed up generation by observing that finding the most violated counterexample is not strictly necessary for enforcement to work.
At the same time, however, complete search is needed (at least once) to determine feasibility -- i.e. to check whether a counterexample exists at all.
We meet these apparently contradicting needs by solving Problem \eqref{eq:generator} with a timeout extension scheme.
Namely:
1) we start the solution process with a timeout;
2) if any counterexample is found or infeasibility is proven, we stop;
3) otherwise, we double the timeout and continue searching;
4) we proceed until the timeout reaches a maximum allowed value, at which point we stop.
In the latter case, since we use Mathematical Programming as our solution technology, we can still provide a bound on the maximum violation level for the network.
The procedure is described in the supplemental material.
The original generator from \cite{francobaldi2025smle} was limited to trace properties and always relied on complete search, which we found to be impractical in our setting.

\input{algos/trainloss}

\paragraph{Training Loss}

During our training phase, we use the multi-component loss function described in \Cref{alg:trainloss}, which incorporates:
1) a term $L_{\text{acc}}$ accounting for model accuracy;
2) a term $L_{\text{box}}$ penalizing box degeneracy, analogously to \Cref{alg:pretrainloss}; 
3) a term $L_{\text{prop}}$ representing the degree of violation for the incumbent counterexample.
At line 11, the multiplier $\lambda_{\text{box}}$ for $L_{\text{box}}$ is fixed, while the one for $L_{\text{prop}}$ is trainable.
During the primal gradient step, we differentiate w.r.t. all $\theta$ parameters and move opposite to the gradient (to reduce the loss value).
During the dual gradient step, we differentiate w.r.t. $\lambda_{\text{box}}$ and move in the same direction as the gradient, to increase the penalty in case of violations.
Results from classical penalty methods ensure that the process asymptotically convergences to a local optimum.

Designing the $L_{\text{prop}}$ term is challenging, as it should not only allow for backpropagation, but also be non-trivial to resolve.
Specifically, in our considered setting we have that, due to the linearity of $g$, for any locally optimal counterexample the two embeddings always lie at one vertex of their bounding boxes: $\forall i \in [n], \; \clipped{z}'_i = \underline{h}(x; \theta_{\underline{h}}) \lor \clipped{z}'_i = \overline{h}(x; \theta_{\overline{h}})_i$, and similarly for $\clipped{z}''$.
In this situation, the counterexample could be resolved by making minor adjustments to the auxiliary models until either $\clipped{z}'$ or $\clipped{z}''$ lie outside the overapproximation box.
In turn, this would lead to very frequent generator calls and impractically large training times.

We address these issues by applying an \emph{abstraction step} to our counterexamples.
Namely, we represent them through the associated active constraints from the overapproximation box, i.e., either $\clipped{z}'_i = \underline{h}(x; \theta_{\underline{h}})_i$ or $\clipped{z}'_i = \overline{h}(x; \theta_{\overline{h}})_i$.
The \textsc{CEProp} routine manages this step and is described in the supplemental material.
An example outcome for the main training loop can be found in \Cref{fig:synthetic_example}: during the dual ascent process, the overapproximation defined by $\underline{h}, \overline{h}$ has expanded, allowing the backbone to more accurately approximate the underlying function in the regions where its shape is compatible with the target constraint (robustness).

The main training loop in \cite{francobaldi2025smle} is based on Projected Gradient Descent (PGD), interleaving primal gradient steps with optimal projection steps limited to the output layer weights.
The latter restriction proved enough to make the approach ineffective in the case of relational properties.
In addition, our method reduces the zig-zag behavior typical of PGD, hence accelerating convergence, and requires significantly fewer generator calls.

\input{algos/projector}

\paragraph{Posttraining}

The main training loop is effective at improving both the accuracy and the degree of property satisfaction of the model.
However, the need to manage those two, often conflicting goals prevents reaching guaranteed feasibility in most cases.
Typically, at this stage only counterexamples associated to modest violation values remain, which we address in the posttraining phase.

For this phase, we use a projection operator implemented via gradient steps, detailed in \Cref{alg:projector}.
Formally, we rely once again on dual ascent, but compared with the main training loop the accuracy term $L_{\text{acc}}$ is not given by a data-driven loss, but rather consists of a squared L2 regularizer on the model weights.
% In the posttraining phase, the \textsc{Generator} and \textsc{Projector} (\Cref{alg:projector}) are executed iteratively until no more couterexample can be found, or the iteration limit is reached.
Upon termination, the loop returns the model weights $\theta$, together with a certified upper bound on property violation \texttt{ViolBound}, whose value depends on the termination condition: if we manage to prove infeasibility in the generator, this value is zero and the model has satisfaction guarantees;
otherwise, the bound can still serve as a form of partial certification.
\Cref{fig:synthetic_example} shows an example outcome for the post-training phase, where the overapproximation box has tightened again to prevent violations.

%% file: tables/properties.tex
\begin{comment}
\begin{table}[tb]
    \renewcommand{\arraystretch}{2}
    \centering
    \begin{tabular}{l|l|l}
         Property & $\underline{\delta}, \overline{\delta}$ & $\underline{\epsilon}, \overline{\epsilon}$\\
         \hline
         Robustness & $\underline{\delta}_i=\overline{\delta}_i=\delta, \forall i \geq 0$&
         $\underline{\epsilon}=\overline{\epsilon}=\epsilon \geq 0$\\
         \hline
         \makecell{Counterfactual \\ Fairness} & \makecell{$\underline{\delta}_i=\overline{\delta}_i=0, \forall i\notin \mathcal{P}$ \\ $\underline{\delta}_i = l_i, \overline{\delta}_i = u_i, \forall i\in \mathcal{P}$} & $\underline{\epsilon}=\overline{\epsilon}=\epsilon \geq 0$\\
         \hline
         \makecell{Non-Increasing \\ Monotonicity} & \makecell{$\underline{\delta}_i=\overline{\delta}_i=0, \forall i\notin \mathcal{P}$ \\ $\underline{\delta}_i = l_i, \overline{\delta}_i = 0, \forall i\in \mathcal{P}$} & $\underline{\epsilon}=0, \overline{\epsilon}=M$ \\
         \hline
        \makecell{Non-decreasing \\ Monotonicity} & \makecell{$\underline{\delta}_i=\overline{\delta}_i=0, \forall i\notin \mathcal{P}$ \\ $\underline{\delta}_i = l_i, \overline{\delta}_i = 0, \forall i\in \mathcal{P}$} & $\underline{\epsilon}=-M, \overline{\epsilon}=0$ 
    \end{tabular}
    \caption{Property derivation}
    \label{tab:properties}
\end{table}
\end{comment}

\begin{table}[tb]
    \renewcommand{\arraystretch}{1.}
    \centering
    \begin{tabular}{ccccccc}\toprule
        \emph{Property} & 
        $\underline{\delta}_{\mathcal{P}}$ & 
        $\underline{\delta}_{\mathcal{P}^c}$ & 
        $\overline{\delta}_{\mathcal{P}}$ & 
        $\overline{\delta}_{\mathcal{P}^c}$ & 
        $\underline{\epsilon}$ & 
        $\overline{\epsilon}$ \\
        \midrule
        Robustness & 
        $\delta$ & 
        $\delta$ & 
        $\delta$ & 
        $\delta$ & 
        $\epsilon$ & 
        $\epsilon$ \\
        %\hline
        \makecell{Fairness} & 
        $l_i$ & 
        $0$ & 
        $u_i$ & 
        $0$ & 
        $\epsilon$ & 
        $\epsilon$ \\
        %\hline
        \makecell{Monotonicity ($\uparrow$)} & 
        $l_i$ & 
        $0$ & 
        $0$ & 
        $0$ & 
        $-M$ & 
        $0$ \\ \bottomrule
        %\hline
        % \makecell{Monotonicity ($\downarrow)$} & 
        % $l_i$ & 
        % $0$ & 
        % $0$ & 
        % $0$ & 
        % $0$ & 
        % $M$ \\ 
    \end{tabular}
    \caption{Configuration of $\delta$ and $\epsilon$ for different properties.}
    \label{tab:properties}
\end{table}

%% file: figures/smile.tex
\begin{figure}[tb]
    \centering
    \includegraphics[width=0.8\linewidth]{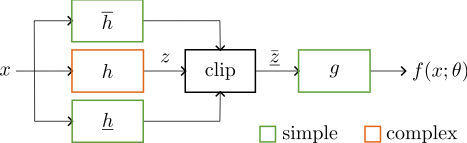}
    \caption{A depiction of the SMiLE architecture.}
    \label{fig:smle_arch}
\end{figure}

%% file: algos/training.tex
\begin{algorithm}[tb]
\caption{\textsc{Train}$(L_{\text{s}}, \theta, \lambda_{\text{box}}, t, x, y, \text{SGD-pars})$}
\label{alg:training}
\begin{algorithmic}[1]
    \STATE \textbf{Pretraining:}
    \FOR{pretraining step}
        \STATE $L \gets \textsc{PretrainLoss}(L_{\text{s}}, \theta, \lambda_{\text{box}}, x, y)$
        \STATE $\theta \gets \textsc{PrimalStep}(L)$
    \ENDFOR
    \vspace{3pt}
    \STATE \textbf{Training:}
    \STATE $x',x'',\clipped{z}',\clipped{z}'', \overline{\gamma}, S \gets \textsc{Generator}(\theta_{\underline{h}}, \theta_{\overline{h}}, \theta_g,t)$
    \STATE $\lambda_{\text{prop}} \gets 0$
    \FOR{training step}
        \IF{$\textsc{Resolved}(x',x'',\clipped{z}',\clipped{z}'')$}
            \STATE $x',x'',\clipped{z}',\clipped{z}'', \overline{\gamma}, S \gets \textsc{Generator}(\theta_{\underline{h}}, \theta_{\overline{h}}, \theta_g,t)$
        \ENDIF
        \STATE $L \gets \textsc{TrainLoss}(L_{\text{s}}, \theta, \lambda_{\text{box}}, \lambda_{\text{prop}}, x, y, x',x'',\clipped{z}',\clipped{z}'')$
        \STATE $\theta \gets \textsc{PrimalStep}(L)$
        \STATE $\lambda_{\text{prop}} \gets \textsc{DualStep}(L)$
    \ENDFOR
    \vspace{3pt}
    \STATE \textbf{Posttraining:}
    \FOR{posttraining step}
        \STATE $x',x'',\clipped{z}',\clipped{z}'', \overline{\gamma}, S \gets \textsc{Generator}(\theta_{\underline{h}}, \theta_{\overline{h}}, \theta_g, t)$
        \IF{$S = \text{infeasible}$}
            \STATE $\texttt{ViolBound} \gets 0$
            \STATE \textbf{break}
        \ENDIF
        \STATE $\theta_{\underline{h}}, \theta_{\overline{h}}, \theta_g \gets \textsc{Projector}(x',x'',\clipped{z}',\clipped{z}'')$
    \ENDFOR
    \STATE $\texttt{ViolBound} \gets \overline{\gamma}$
    \RETURN $\theta, \texttt{ViolBound}$
\end{algorithmic}
\end{algorithm}

\begin{comment}
\begin{algorithm}[tb]
\caption{\textsc{Trainer}$(\textit{train-data}, \textit{SGD-params}, \theta)$}
\label{alg:projector}
\begin{algorithmic}
    \STATE \textbf{Pretraining:}
    \FOR{pretraining step}
        \STATE compute $L_{\text{pretr}}(\theta)$ as in \cref{eq:loss_pretr}
        \STATE update $\theta$ by descending $\nabla_\theta L_{\text{pretr}}(\theta)$ 
    \ENDFOR
    \vspace{3pt}
    \STATE \textbf{Training:}
    \FOR{training step}
        \STATE $(x',x'',\clipped{z}',\clipped{z}'') \gets$ \textsc{Generator}$(\theta_{\underline{h}}, \theta_{\overline{h}}, \theta_g,t)$
        \STATE compute $L_{\text{tr}}(\theta, x',x'',\clipped{z}',\clipped{z}'')$ as in \cref{eq:loss_tr}
        \STATE update $\theta$ by descending $\nabla_\theta L_{\text{tr}}(\theta, \lambda, x',x'',\clipped{z}',\clipped{z}'')$
        \STATE update $\lambda$ by ascending $\nabla_\lambda L_{\text{tr}}(\theta, \lambda, x',x'',\clipped{z}',\clipped{z}'')$
    \ENDFOR
    \vspace{3pt}
    \STATE \textbf{Posttraining:}
    \FOR{posttraining step}
        \STATE $(x',x'',\clipped{z}',\clipped{z}'') \gets \textsc{Generator}(\theta_{\underline{h}}, \theta_{\overline{h}}, \theta_g, t)$
        \STATE $L_{\text{prop}} \gets \max(0, \max(\underline{\varepsilon} - y' + y'', y' - y'' - \bar{\varepsilon}))$
        \IF{$L_{\text{prop}} = 0$}
            \STATE \textbf{Stop}
        \ENDIF
        \STATE $\theta_{\underline{h}}, \theta_{\overline{h}}, \theta_g \gets \textsc{Projector}(x',x'',\clipped{z}',\clipped{z}'')$
    \ENDFOR
\end{algorithmic}
\end{algorithm}
\end{comment}

%% file: figures/synthetic_example.tex
\begin{figure}[tb]
    \centering
    \includegraphics[width=0.7\linewidth]{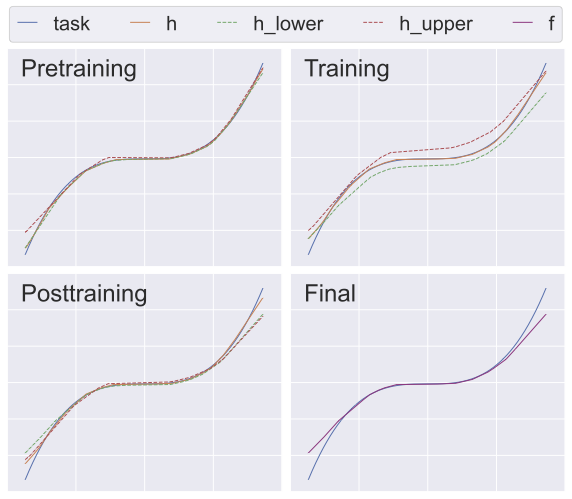}
    \caption{A SMiLE model $f$ with latent dimension $n=1$ and output function $g = \text{id}$, trained on the task $y = x^3$, and enforced with $|x' - x''| \leq 0.1 \implies |f(x') - f(x'')| \leq 0.5\max_{x \in \mathcal{X}, \delta \leq 0.1} |x^3 - (x + \delta')^3|$, i.e., we want the model to be $0.5$ times more robust than the task itself. The subplots show the evolution of the model during its training.}
    \label{fig:synthetic_example}
\end{figure}

%% file: algos/pretrainloss.tex
\begin{algorithm}[tb]
\caption{\textsc{PretrainLoss}$(L_{\text{s}}, \theta, \lambda_{\text{box}}, x, y)$}
\label{alg:pretrainloss}
\begin{algorithmic}[1]
    \STATE $L_f \gets L_{\text{s}}(y, f(x;\theta))$
    \STATE $L_{\underline{h}} \gets L_{\text{s}}(y, g(\underline{z};\theta_g) \circ \underline{h}(x;\theta_{\underline{h}}))$
    \STATE $L_{\overline{h}} \gets L_{\text{s}}(y, g(\overline{z};\theta_g) \circ \overline{h}(x;\theta_{\overline{h}}))$
    \STATE $L_h \gets L_{\text{s}}(y, g(z;\theta_g) \circ h(x;\theta_h))$
    \STATE $L_{\overline{\text{box}}} \gets \max(0, h(x;\theta_h) - \overline{h}(x,\theta_{\overline{h}})) $
    \STATE $L_{\underline{\text{box}}} \gets \max(0, \underline{h}(x;\theta_{\underline{h}}) - h(x;\theta_h))$
    \STATE $L_{\text{acc}} \gets L_f + L_{\underline{h}} + L_{\overline{h}} + L_h$, \; $L_{\text{box}} \gets L_{\overline{\text{box}}} + L_{\underline{\text{box}}}$
    \RETURN $L_{\text{acc}} + \lambda_{\text{box}} L_{\text{box}}$ 
\end{algorithmic}
\end{algorithm}

%% file: algos/trainloss.tex
\begin{algorithm}[tb]
\caption{\textsc{TrainLoss}$(L_{\text{s}}, \theta, \lambda_{\text{box}}, \lambda_{\text{prop}}, x, y, x',x'',\clipped{z}',\clipped{z}'')$}
\label{alg:trainloss}
\begin{algorithmic}[1]
    \STATE $L_{\text{acc}} \gets L_{\text{s}}(y, f(x;\theta))$
    \STATE $L_{\overline{\text{box}}} \gets \max(0, h(x;\theta_h) - \overline{h}(x,\theta_{\overline{h}})) $
    \STATE $L_{\underline{\text{box}}} \gets \max(0, \underline{h}(x;\theta_{\underline{h}}) - h(x;\theta_h))$
    \STATE $L_{\text{box}} \gets L_{\overline{\text{box}}} + L_{\underline{\text{box}}}$
    \IF{$(x',x'',\clipped{z}',\clipped{z}'') \neq \text{null}$}
        \STATE $y' \gets \textsc{CEProp}(\theta_{\underline{h}}, \theta_{\overline{h}}, \theta_g, x', \clipped{z}')$
        \STATE $y'' \gets \textsc{CEProp}(\theta_{\underline{h}}, \theta_{\overline{h}}, \theta_g, x'', \clipped{z}'')$
        \STATE $L_{\text{prop}} \gets \max(0, \max(\underline{\varepsilon} - y' + y'', y' - y'' - \bar{\varepsilon}))$
    \ELSE
        \STATE $L_{\text{prop}} \gets 0$
    \ENDIF
        \RETURN $L_{\text{acc}} + \lambda_{\text{box}} L_{\text{box}} + \lambda_{\text{prop}} L_{\text{prop}}$ 
\end{algorithmic}
\end{algorithm}

%% file: algos/projector.tex
\begin{algorithm}[tb]
\caption{\textsc{Projector}$(x', x'', \clipped{z}', \clipped{z}'', \theta_{\underline{h}}, \theta_{\overline{h}}, \theta_g, \text{SGD-pars})$}
\label{alg:projector}
\begin{algorithmic}[1]
\STATE $\theta_{\underline{h}}^{\text{orig}} \gets \theta_{\underline{h}}, \; \theta_{\overline{h}}^{\text{orig}} \gets \theta_{\overline{h}}, \; \theta_g^{\text{orig}} \gets \theta_g, \lambda_{\text{prop}} \gets 0$
\FOR{projection step}
    \STATE $y' \gets \textsc{CEProp}(\theta_{\underline{h}}, \theta_{\overline{h}}, \theta_g, x', \clipped{z}')$
    \STATE $y'' \gets \textsc{CEProp}(\theta_{\underline{h}}, \theta_{\overline{h}}, \theta_g, x'', \clipped{z}'')$
    \STATE $L_{\text{acc}} \gets \|\theta_{\underline{h}}^{\text{orig}} - \theta_{\underline{h}}\|_2^2 + \|\theta_{\overline{h}}^{\text{orig}} - \theta_{\overline{h}}\|_2^2 + \|\theta_{g}^{\text{orig}} - \theta_{g}\|_2^2$
    \STATE $L_{\text{prop}} \gets \max(0, \max(\underline{\varepsilon} - y' + y'', y' - y'' - \bar{\varepsilon}))$
    \STATE $L \gets L_{\text{acc}} + \lambda_{\text{prop}} L_{\text{prop}}$
    \STATE $\theta_{\underline{h}}, \theta_{\overline{h}}, \theta_g \gets \textsc{PrimalStep}(L)$
    \STATE $\lambda_{\text{prop}} \gets \textsc{DualStep}(L)$
    \ENDFOR
\RETURN $\theta_{\underline{h}}, \theta_{\overline{h}}, \theta_g$
\end{algorithmic}
\end{algorithm}

%% file: experimentation.tex
We evaluate our framework on three benchmarks: \emph{monotonicity}, \emph{fairness} and \emph{robustness}, guided by the following research questions:
\emph{(Q1: Guarantees)} Can SMiLE consistently provide property satisfaction guarantees?  
\emph{(Q2: Accuracy)} Is the accuracy of SMiLE competitive with state-of-the-art property-specific methods? 
\emph{(Q3: Applicability)} Can SMiLE provide practically valuable outcomes?
\emph{(Q4: Generality)} Can SMiLE consistently enforce different properties on different neural architectures?

Our experiments are developed in Python: we implemented SMiLE by using Keras \cite{chollet2015keras}, Pyomo \cite{hart2011pyomo, bynum2021pyomo}, OMLT \cite{ceccon2022omlt} and Gurobi \cite{gurobi}, while for the considered competitors we used the official code released by their authors. 

\paragraph{Monotonicity}
We consider 9 synthetic tasks given by the function $y = x + \alpha \sin(\omega x)$, for $\alpha \in \{2, 3, 4\}$ and $\omega \in \{0.4, 0.6, 0.8\}$, which becomes increasingly non-monotonic as $\alpha$ and $\omega$ increase. 
On each task, we train a SMiLE model by enforcing non-decreasing monotonicity. Precisely, for any input pair $x',x''$ such that $x' \leq x''$, we force the corresponding outputs to satisfy $f(x') \leq f(x'')$.
\Cref{fig:monotonicity} depicts a heatmap with the $R^2$ performance of our model (the higher, the better), and an example of a monotonic approximation of a non-monotonic function, showing how our method can achieve acceptable results even when the property is substantially violated in the data. 
In all cases, 0-violation was reached after posttraining.
% We highlight, in particular, that the training of each model converged to zero property violation.
\input{figures/monotonicity}
\paragraph{Robustness} We consider the task of detecting the digit ``zero'' (binary classification) on the MNIST dataset \cite{lecun1998mnist}.
We train 15 SMiLE models, each time by enforcing, on the model logit, a $\delta,\epsilon$-robustness property with $\delta \in \{0.010, 0.025, 0.050, 0.075, 0.100\}$ and $\epsilon \in \{0.75, 1.00, 1.25\}$. Precisely, for any input pair $x',x''$ such that $\|x'-x''\|_{\infty} \leq \delta$, we force the corresponding logits  to satisfy $|f_{\text{logit}}(x') - f_{\text{logit}}(x'')| \leq \epsilon$.

The global robustness guarantees from our method can be used to implement a \emph{constant-time} rejection-based defense against adversarial attacks.
The defense checks whether the logit $f_{\text{logit}}(x)$, for a test input $x$, lies outside the rejection region $[-\epsilon, \epsilon]$.
In this case, it issues a safety certificate: even if $x$ was adversarially generated from a clean input as specified in the property,
% , generated within an $l_\infty$ perturbation radius $\delta$ from a clean input $\bar{x}$,
the attack would be unable to change the logit by more than $\epsilon$, and hence to flip the prediction.
In the opposite case, the defense raises a warning to the user.

We evaluate our framework against CROWN-IBP \cite{huan2020towards}, which penalizes property violation by training the model against a convex combination of a standard and a robustness loss, computed via a sound but incomplete robustness verifier, integrating CROWN \cite{zhang2018efficient} with IBP \cite{gowal2019effectiveness}. 
The method depends on both the perturbation $\delta$ and a convex combination hyperparameter $\lambda$, dynamically adjusted according to a scheduling strategy.
At training time, CROWN-IBP enforces robustness only locally (i.e., around the training samples); 
at inference time, it provides a defense mechanism similar to the one described above, except that the rejection region needs to be computed for each sample by solving an overapproximation.
% similarly to us, it either certifies the non-existence of an attack within the enforced $\delta$ for a given input $x$, or fails to provide such certificate.
We train CROWN-IBP for the same values of $\delta$ considered for SMiLE. Together with it, we also consider a base model unaware of the property to satisfy (Agnostic). 
In all approaches, we adopt a \emph{deep convolutional neural network} for $h$, and \emph{linear models} for $\underline{h}, \overline{h}$.

\input{tables/robustness_accuracy}

The predictive performance of the three competitors on the first 1,000 instances of the MNIST dataset is reported in \Cref{tab:robustness_accuracy}, where the SMiLE results are aggregated across the different $\epsilon$, and where \emph{Clean}, \emph{PGD} and \emph{Verified} denote the percentage of correct predictions on the unperturbed test set, on the test set under a 100-step PGD attack, and the percentage of correct and verified predictions under any attack, respectively.
We highlight, moreover, that all reported SMiLE models are guaranteed robust: their training successfully terminated with zero property violation, enabling the corresponding real-time defense.
CROWN-IBP emerges as the most accurate model across all metrics. Agnostic maintains constant clean accuracy and zero verified one, being unaffected by $\delta$ and unable to certify, while its PGD accuracy drops significantly under stronger attacks, highlighting its vulnerability. SMiLE, on the other hand, while less accurate than the property-specific CROWN-IBP, exhibits a moderate decline in clean, verified, and PGD accuracy as attacks intensify, demonstrating its ability to produce verifiable and also significantly more robust networks than a baseline approach.
Finally, while SMiLE may not match CROWN-IBP in terms of accuracy, it clearly outperforms it in terms of runtime: we report an average training time of 14,138s for CROWN-IBP, and of 7,717s for SMiLE; more importantly, at inference time, CROWN-IBP requires a total of 11.20s to both predict and verify on the test set, while SMiLE employs only 0.03s for the same procedure, comparable with the base model, which predicts in 0.02s but without certifying.

\paragraph{Fairness}
We consider 3 widely used datasets in the fairness literature: Compas \cite{angwin2016compas}, Law \cite{wightman1998law} and Crime \cite{redmond2002communities}, where the task is to predict recidivism, law school admission and crime rate, respectively.
The first two are binary classification tasks, while the third is a regression one.
On each dataset, we train 5 SMiLE models, each time by enforcing, on the model output (logit), an $\epsilon$-fairness property, with $\delta = 0$ and $\epsilon \in \{0.2, 0.4, 0.6, 0.8, 1.0\}$ for Compas, $\epsilon \in \{1, 2, 3, 4, 5\}$ for Law and $\epsilon \in \{0.02, 0.04, 0.06, 0.08, 0.10\}$ for Crime. Precisely, for any input pair $x',x''$ such that $|x'_i-x''_i| \leq \delta \; \forall i \neq p$ (i.e., $x',x''$ are mutually counterfactual), we force the corresponding outputs (logits) to satisfy $|f(x') - f(x'')| \leq \epsilon$, where the protected variable $p$ is always \emph{race}. 
We evaluate our framework against CertiFair \cite{khedr2023certifair}, which penalizes property violation by training the model against a convex combination of a standard and a fairness loss, computed via a sound but incomplete fairness verifier.
The method depends on a combination hyperparameter $\lambda$, while is independent of the counterfactual perturbation $\delta$.
At training time, similarly to us, our competitor enforces fairness globally (i.e., on the entire input space); at inference time, differently from us, it lacks property satisfaction guarantees on individual inputs.
We train CertiFair for 5 values of $\lambda$, calibrated to span the reasonable hyperparameter space for each dataset: $\lambda \in \{0.025, 0.050, 0.075, 0.100, 0.125\}$ for Compas and Law, and $\lambda \in \{0.08, 0.09, 0.10, 0.11, 0.12\}$ for Crime.
Together with it, we consider again Agnostic.
In all approaches, we adopt a \emph{deep feedforward neural network} for $h$, and 1-hidden-layer \emph{ReLU neural networks} for $\underline{h}, \overline{h}$.
\input{figures/fairness}
\input{figures/counter_var}

We report the results of the experiment in \Cref{fig:fairness}, where each dot represents a model evaluated along two dimensions, \emph{Predictive Quality} and \emph{Counterfactual Variation}: the former corresponds to $R^2$ or accuracy (the higher, the better), the latter to the maximum absolute difference in model output (logit) on any pair of counterfactual samples from the test set, quantifying unfairness (the lower, the better), where the counterfactual of a sample is obtained by flipping its protected attribute.
The figure clearly demonstrates the superiority of our method across all datasets: SMiLE models are consistently positioned to the left of Agnostic models, indicating a substantially higher degree of fairness compared to the baseline approach, as well as above and generally to the left of CertiFair models, showing that our method achieves higher accuracy while guaranteeing an equivalent or superior level of fairness relative to its competitor.
We highlight that the reported SMiLE models achieve again zero property violation, meaning that besides decreasing the counterfactual variation in-distribution (on the test set), we are also able to guarantee a variation upper bound (i.e., the enforced $\epsilon$) out of distribution.
Finally, \Cref{fig:counter_var} shows how the counterfactual variation of SMiLE (color gradient) decreases with $\epsilon$, over the pair of counterfactual samples (dots) from the Crime test set, projected on two non-protected features.

\paragraph{Discussion}
\emph{(Q1: Guarantees)} Each SMiLE model in our computational study successfully converged to zero property violation, demonstrating that, in practice, our framework can consistently provide satisfaction guarantees.
\emph{(Q2: Accuracy)} While SMiLE is outperformed by CROWN-IBP on \emph{robustness}, it shows high accuracy on \emph{monotonicity} and \emph{fairness}, where it matches Agnostic and surpasses CertiFair, highlighting its competitive predictive capabilities.
\emph{(Q3: Applicability)} On \emph{robustness} and \emph{fairness}, SMiLE is able to produce more robust and fair networks, as well as to enable a very fast real-time defense against adversarial attacks, yielding practically valuable outcomes.
\emph{(Q4: Generality)} The strong results achieved across three relational properties (monotonicity, robustness, and fairness), and for different neural architectures (convolutional and feedforward), demonstrate the broad generality of SMiLE.

%% file: figures/monotonicity.tex
\begin{figure}[tb]
    \centering
    \includegraphics[width=0.49\linewidth]{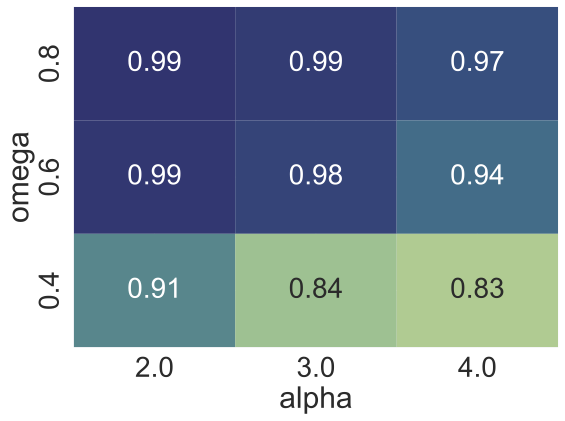}
    \includegraphics[width=0.49\linewidth]{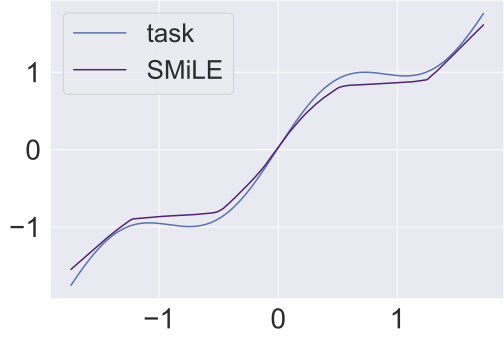}
    \caption{Accuracy \emph{monotonicity} (left) and an example of monotonic estimator for the non-monotonic function given by $\alpha = 2$ and $\omega = 0.6$ (right).}
    \label{fig:monotonicity}
\end{figure}

%% file: tables/robustness_accuracy.tex
\begin{table}[tb]
    \renewcommand{\arraystretch}{1.}
    \centering
    \begin{tabular}{ccrrr}\toprule
        $\mathbf{\delta}$&\emph{Model}&\emph{Clean}&\emph{PGD}&\emph{Verified}\\\midrule
        0.010&Agnostic&99.60&99.50&00.00\\
        0.010&CROWNIBP&100.00&99.90&99.90\\
        0.010&SMiLE&98.80&98.47&97.57\\
        \midrule
        0.025&Agnostic&99.60&98.40&00.00\\
        0.025&CROWNIBP&99.80&99.70&99.60\\
        0.025&SMiLE&98.60&97.90&96.37\\
        \midrule
        0.050&Agnostic&99.60&90.40&00.00\\
        0.050&CROWNIBP&99.70&99.40&99.20\\
        0.050&SMiLE&97.87&96.07&94.33\\
        \midrule
        0.075&Agnostic&99.60&75.70&00.00\\
        0.075&CROWNIBP&99.60&99.20&99.20\\
        0.075&SMiLE&97.07&94.20&92.37\\
        \midrule
        0.100&Agnostic&99.60&66.80&00.00\\
        0.100&CROWNIBP&99.60&99.30&99.00\\
        0.100&SMiLE&95.87&92.17&90.57 \\ \bottomrule
    \end{tabular}
    \caption{Accuracy on \emph{robustness}.} 
    \label{tab:robustness_accuracy}
\end{table}

%% file: figures/fairness.tex
\begin{figure}[tb]
    \centering
    \includegraphics[width=\linewidth]{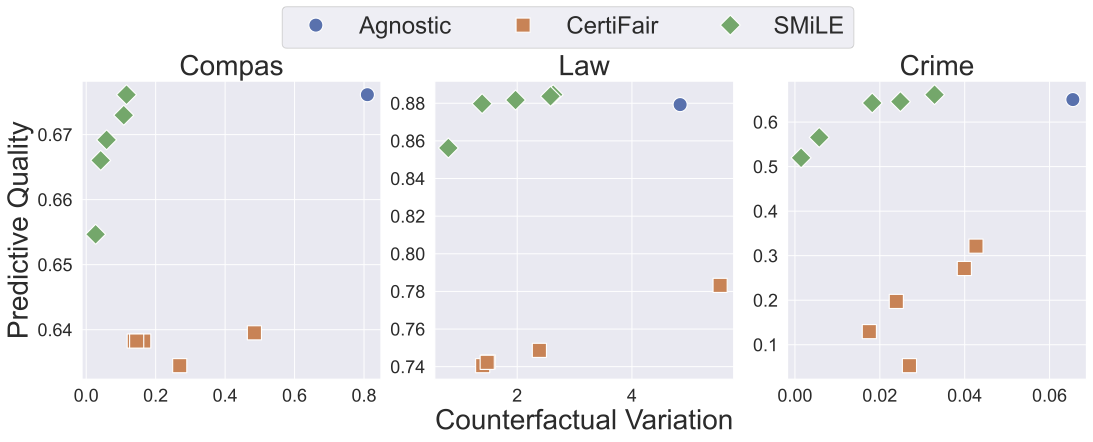}
    \caption{Accuracy and counterfactual variation on \emph{fairness}.}
    \label{fig:fairness}
\end{figure}

%% file: figures/counter_var.tex
\begin{figure}[tb]
    \centering
    \includegraphics[width=\linewidth]{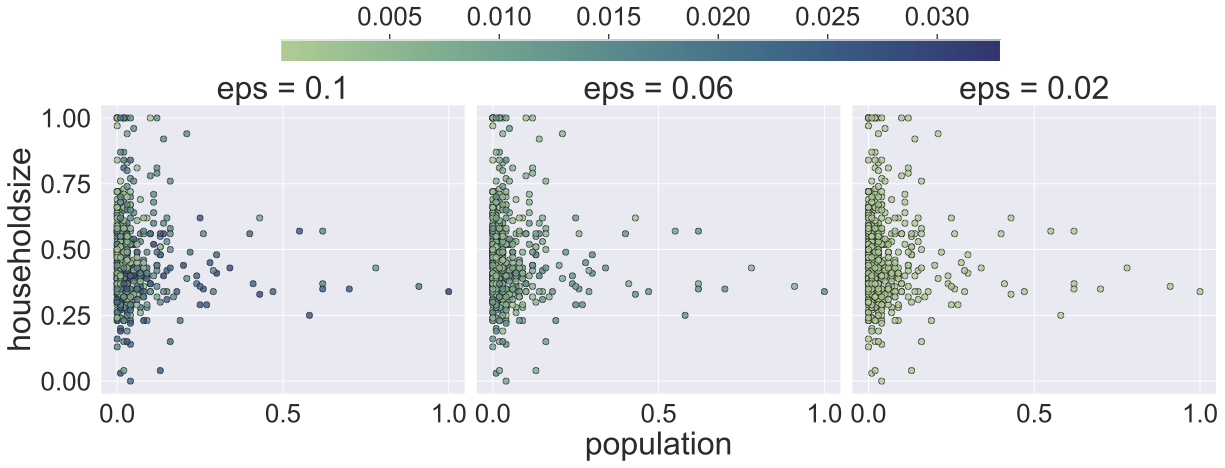}
    \caption{Counterfactual variation for different $\epsilon$ on
Crime.}
    \label{fig:counter_var}
\end{figure}

%% file: conclusion.tex
We extended the property-enforcement SMiLE framework, designed for trace properties, to also support relational ones, by retaining its verification-friendly architecture, while completely reengineering its training algorithm.
Precisely, we designed a procedure consisting of 3 phases: \emph{pretraining}, which suitably initializes the model to prevent poor local optima; \emph{training}, which maximizes accuracy while encouraging feasibility via a counterexample-based Dual Gradient Ascent; and \emph{posttraining}, which eliminates residual violations by iteratively generating and resolving counterexamples.  
We tested our framework on \emph{monotonicity}, \emph{robustness} and \emph{fairness}, for regression and classification tasks, demonstrating its ability to provide satisfaction guarantees for generic relational properties and networks, while remaining competitive with property-specific methods.
Our contributions open up several research directions, such as further extending the SMiLE framework to \emph{functional properties}, constraining input and output together into the same predicates, or adapting it to inputs of variable size, via Transformers or Graph Neural Networks.

%% file: reproducibility_checklist.tex
Unless specified otherwise, please answer ``yes'' to each question if the relevant information is described either in the paper itself or in a technical appendix with an explicit reference from the main paper. If you wish to explain an answer further, please do so in a section titled ``Reproducibility Checklist'' at the end of the technical appendix.

\subsection{General Paper Structure}
\begin{enumerate}
    \item Includes a conceptual outline and/or pseudocode description of AI methods introduced (yes/partial/no/NA) -- Yes
    \item Clearly delineates statements that are opinions, hypothesis, and speculation from objective facts and results (yes/no) -- Yes
    \item Provides well marked pedagogical references for less-familiare readers to gain background necessary to replicate the paper (yes/no) -- Yes
\end{enumerate}

\subsection{Theoretical Contributions}
\begin{enumerate}
    \item Does this paper make theoretical contributions? (yes/no) Yes \\
    If yes, please complete the list below.\\
    \item All assumptions and restrictions are stated clearly and formally. (yes/partial/no) -- Yes
    \item All novel claims are stated formally (e.g., in theorem statements). (yes/partial/no) -- Partial
    \item Proofs of all novel claims are included. (yes/partial/no) -- No
    \item Proof sketches or intuitions are given for complex and/or novel results. (yes/partial/no) -- Yes
    \item Appropriate citations to theoretical tools used are given. (yes/partial/no) -- Yes
    \item All theoretical claims are demonstrated empirically to hold. (yes/partial/no/NA) -- Yes
    \item All experimental code used to eliminate or disprove claims is included. (yes/no/NA) -- Yes
\end{enumerate}

\subsection{Dataset Usage}
\begin{enumerate}
    \item Does this paper rely on one or more datasets? (yes/no) -- Yes \\ 
    If yes, please complete the list below.\\

    \item A motivation is given for why the experiments are conducted on the selected datasets (yes/partial/no/NA) -- Yes
    \item All novel datasets introduced in this paper are included in a data appendix. (yes/partial/no/NA) -- Yes
    \item All novel datasets introduced in this paper will be made publicly available upon publication of the paper with a license that allows free usage for research purposes. (yes/partial/no/NA) -- Yes
    \item All datasets drawn from the existing literature (potentially including authors’ own previously published work) are accompanied by appropriate citations. (yes/no/NA) -- Yes
    \item All datasets drawn from the existing literature (potentially including authors’ own previously published work) are publicly available. (yes/partial/no/NA) -- Yes
    \item All datasets that are not publicly available are described in detail, with explanation why publicly available alternatives are not scientifically satisficing. (yes/partial/no/NA) -- NA
\end{enumerate}

\subsection{Computational Experiments}

\begin{enumerate}
    \item Does this paper include computational experiments? (yes/no) -- Yes\\
    If yes, please complete the list below.\\

    \item This paper states the number and range of values tried per (hyper-) parameter during development of the paper, along with the criterion used for selecting the final parameter setting. (yes/partial/no/NA) -- Yes
    \item Any code required for pre-processing data is included in the appendix. (yes/partial/no). -- Yes
    \item All source code required for conducting and analyzing the experiments is included in a code appendix. (yes/partial/no) -- Yes
    \item All source code required for conducting and analyzing the experiments will be made publicly available upon publication of the paper with a license that allows free usage for research purposes. (yes/partial/no) -- Yes
    \item All source code implementing new methods have comments detailing the implementation, with references to the paper where each step comes from (yes/partial/no) -- Yes
    \item If an algorithm depends on randomness, then the method used for setting seeds is described in a way sufficient to allow replication of results. (yes/partial/no/NA) -- Partial
    \item This paper specifies the computing infrastructure used for running experiments (hardware and software), including GPU/CPU models; amount of memory; operating system; names and versions of relevant software libraries and frameworks. (yes/partial/no) -- Yes
    \item This paper formally describes evaluation metrics used and explains the motivation for choosing these metrics. (yes/partial/no) -- Yes
    \item This paper states the number of algorithm runs used to compute each reported result. (yes/no) -- Yes
    \item Analysis of experiments goes beyond single-dimensional summaries of performance (e.g., average; median) to include measures of variation, confidence, or other distributional information. (yes/no) -- Yes
    \item The significance of any improvement or decrease in performance is judged using appropriate statistical tests (e.g., Wilcoxon signed-rank). (yes/partial/no) -- No
    \item This paper lists all final (hyper-)parameters used for each model/algorithm in the paper’s experiments. (yes/partial/no/NA) -- Yes
\end{enumerate}

%% file: supplemental.tex
\section{-- Supplemental Material --}
\section{Trace vs Relational SMiLE}
Even though the original SMiLE framework, designed exclusively for trace properties, readily applies to relational ones, in this setting it proves completely inadequate, due to structural limitations arising from its core components, which we describe below.

\paragraph{Projector}
The original SMiLE framework resolves counterexamples by acting solely on $g$, hence leaving the auxiliaries mostly static, which proves completely insufficient for relational properties. 
In fact, under the influence of the accuracy-driven loss, without an auxiliary-aware projector, the overapproximation tends to remain large, to allow the more expressive backbone to freely operate in between the bounds.
For relational properties this leads to highly violating counterexamples, obtained by picking the embeddings at the opposite corners of their boxes, whose resolution makes $g$ collapse to an almost constant function, being this the only way to ensure that such remote embeddings are propagated within the output bounds. 
One way to prevent such degenerate cases is to incorporate $\underline{h}$ and $\overline{h}$ into the projection step, thereby distributing the burden of the enforcement also onto these components, rather than relying solely on $g$. 
However, this introduces non-linearities and makes exact resolution impractical. 
We thus resort to Dual Ascent, by applying an abstraction step to the embeddings to enable backpropagation, as outlined in \Cref{alg:ceprop}.

\paragraph{Pretraining} 
The purely accuracy-driven pretraining used in trace SMiLE tends to produce large boxes, that is, to move the auxiliary models apart, in order to allow the backbone to freely operate in between them.
Such boxes, moreover, might be non-homogeneous, that is, wide in some regions, tight in others.
While irrelevant in the trace enforcement, where the projector is auxiliary-agnostic, in the relational one the auxiliary-aware projector, to tighten the box in wide regions, where high violations arise, often causes degeneracy (i.e., bound flipping) in already narrow ones. 
When this happens, the semantic of the clipping operator prevents most gradient components from being backpropagated, causing catastrophic training failure. 
To homogenize the initial overapproximation, we warm-start the algorithm through the pretrainig step described in the main paper.

\paragraph{Generator} 
The trace SMiLE framework generates optimal counterexamples, which becomes too costly for relational properties. 
This is why, in the new setting, we adopt a timeout scheme, ensuring acceptable suboptimality in training, while retaining optimality in posttraining to provide guarantees, as described in the main paper.

\section{Additional Algorithms}
Few minor subroutines employed in the SMiLE training framework are only partially described in the main paper due to space limits; precisely: \textsc{Generator}, \textsc{CEProp} and  \textsc{Resolved}.

\textsc{Generator} generates counterexamples to compute property violation, by iteratively solves a MIP problem according to a timeout extension scheme. 
The pseudocode of the timeout criterion is provided in \Cref{alg:generator}, while the underlying MIP model is formulated in \Cref{eq:generator}.

\begin{subequations}
\label{eq:generator}
\begin{align}
    & \Argmax_{x', x'', \overline{\underline{z}}', \overline{\underline{z}}''} \;\gamma \\
    \text{s.t. } 
    & l \leq x', x'' \leq u \\ 
    & \underline{\delta} \leq x' - x'' \leq \overline{\delta} \\
    & \clipped{z}' \geq \underline{h}(x'; \theta_{\underline{h}}) \label{eq:generator_clip_1a}\\
    & \clipped{z}' \leq \underline{h}(x'; \theta_{\underline{h}}) + M (1 - t') \label{eq:generator_clip_1b}\\
    & \clipped{z}' \leq \overline{h}(x'; \theta_{\bar{h}}) + M t' \label{eq:generator_clip_1c}\\
    & \clipped{z}'' \geq \underline{h}(x''; \theta_{\underline{h}}) \label{eq:generator_clip_2a}\\
    & \clipped{z}'' \leq \underline{h}(x''; \theta_{\underline{h}}) + M (1 - t'') \label{eq:generator_clip_2b}\\
    & \clipped{z}'' \leq \overline{h}(x''; \theta_{\bar{h}}) + M t''\label{eq:generator_clip_2c}\\
    & y' = g(\overline{\underline{z}}';\theta_g) \\
    & y'' = g(\overline{\underline{z}}'';\theta_g) \\
    & \gamma \leq y' - y'' -  \overline{\epsilon} + Mb \\
    & \gamma \leq -y' + y'' + \underline{\epsilon} + M(1-b) \\
    & x', x'' \in \mathbb{R}^{B^c} \times \{0,1\}^B, \clipped{z}', \clipped{z}'' \in \mathbb{R}^n, \; y', y'' \in \mathbb{R} \\
    & b \in \{0,1\}, t \in \{0,1\}^n, \; \gamma \in \mathbb{R}_{>0}
\end{align}
\end{subequations}
where $M \geq 0$ is a fixed big-$M$ value, while $B,B^c \subseteq [m]$ respectively denote the set of binary variables and its complement, necessary when the learning task involves binary or one-hot encoded features.
\Cref{eq:generator_clip_1a,eq:generator_clip_1b,eq:generator_clip_1c} and \cref{eq:generator_clip_2a,eq:generator_clip_2b,eq:generator_clip_2c} linearize the constraints
\[
\underline{h}(x; \theta_{\underline{h}}) \leq \clipped{z} \leq \max(\underline{h}(x; \theta_{\underline{h}}), \overline{h}(x; \theta_{\bar{h}})) 
\]
used in the main paper to model bounding box degeneracy, occurring when $\underline{h}(x; \theta_{\underline{h}})_i > \overline{h}(x; \theta_{\bar{h}})_i$ for some $i=1\dots n$.
\input{algos/generator}

\textsc{CEProp} applies an \emph{abstraction step} to a counterexample, by representing it through the combination of the associated active constraints from the overapproximation box, i.e. either $\clipped{z}'_i = \underline{h}(x; \theta_{\underline{h}})_i$ or $\clipped{z}'_i = \overline{h}(x; \theta_{\overline{h}})_i$. 
This routine is necessary to avoid trivial resolutions, obtainable by simply making minor adjustments to the auxiliary models until either $\clipped{z}'$ or $\clipped{z}''$ lies outside the overapproximation box, and is fromalized in {\Cref{alg:ceprop}.
\input{algos/ceprop}

Finally, \textsc{Resolved} checks the existence or validity of a counterexample, in order to eventually trigger the generation of a new one. 
This routine is described in \Cref{alg:resolved}.
\input{algos/resolved}

\section{Experimental Setup}
\paragraph{Hardware Specifications} The software tools used to implement all the considered models, benchmarks and experiments are specified in the main paper. The hardware infrastructure where all the experiments are executed, instead, consists of an Apple M3 Pro CPU with 11 cores and an Apple M3 Pro GPU with 14 cores, equipped with 36 GB RAM and running macOS v15.5 as operating system.

\paragraph{Data Preprocessing}
\emph{Monotonicity} data, i.e., synthetic, is unifomly generated and standardized; \emph{robustness} data, i.e., MNIST, is normalized in between [0,1]; \emph{fairness} data, i.e. Compas, Crime and Law, is preprocessed by dropping any missing rows or columns, normalizing numerical features in between [0, 1], and one-hot encoding categorical ones. 

\paragraph{Model Prametrization}
The main model hyperparameters in our computational study are represented by the backbone architecture $h$ and output function $g$, affecting SMiLE and its competitors (Agnostic, CROWN-IBP and CertiFair), and by the auxiliary models $\underline{h}$ and $\overline{h}$, impacting solely SMiLE.
These key design choices are tailored to the different benchmarks. 
The complete set of architectures used in our experiments is shown in \Cref{tab:architectures}, where $\text{relu}_k$ and $\text{linear}_k$ denote ReLU and linear layers with $k$ neurons, respectively, while $\text{conv}_k$ indicates a convolutional layer with $k$ filters, kernel size $3{\times}3$, stride $(2, 2)$, and ReLU activation.

\input{tables/architectures}
\input{tables/training_hyperparameters}
\input{tables/training_hyperparameters_fairness}

As in any standard Deep Learning setting, our study also requires the selection of the hyperparameters controlling the training processes. 
These hyperparameters, detailed in \Cref{tab:training_hyperparameters,tab:training_hyperparameters_fairness}, were determined based on pilot experiments, aiming at maximizing predictive performance for SMiLE and its competitors.

Finally, beyond the model and training hyperparameters, our computational study involves an additional set of minor, less critical decisions (e.g. random seeds for randomness-affected routines, training initialization, etc), which can be obtained by inspecting the code provided in the Code \& Data Annex.  

%% file: algos/generator.tex
\begin{algorithm}[tb]
\caption{\textsc{Generator}$(\theta_{\underline{h}}, \theta_{\overline{h}}, \theta_g, t)$}
\label{alg:generator}
\begin{algorithmic}[1]
    \STATE $(x', x'', \clipped{z}', \clipped{z}'') \gets \text{null}$
    \FOR{each generation step}
        \STATE $(x', x'', \clipped{z}', \clipped{z}'')$ resume solve \cref{eq:generator} with limit $t$
        \STATE S $\gets$ solver termination status, \; , $\overline{\gamma} \gets$ best dual bound
        \IF{$(x', x'', \clipped{z}', \clipped{z}'') \neq \text{null}$}
            \STATE \textbf{break}
        \ENDIF
        \IF{S $=$ infeasible}
            \STATE \textbf{break}
        \ENDIF
    \STATE $t \gets 2 \cdot t$
    \ENDFOR
    \RETURN $x', x'', \clipped{z}', \clipped{z}'', \overline{\gamma}, S$
\end{algorithmic}
\end{algorithm}

%% file: algos/ceprop.tex
\begin{algorithm}[tb]
\caption{\textsc{CEProp}$(\theta_{\underline{h}}, \theta_{\overline{h}}, \theta_g, x, \clipped{z})$}
\label{alg:ceprop}
\begin{algorithmic}[1]
\STATE $\underline{I} \gets \{i\colon \clipped{z}_i = \underline{h}(x;\theta_{\underline{h}})_i\}, \; \overline{I} \gets \overline{I}^c$
\STATE $\clipped{z} = (\underline{h}(x;\theta_{\underline{h}})_{\underline{I}}, \overline{h}(x;\theta_{\overline{h}})_{\overline{I}}), \; y = g(\clipped{z}, \theta_g)$
\RETURN $y$
\end{algorithmic}
\end{algorithm}

%% file: algos/resolved.tex
\begin{algorithm}[tb]
\caption{\textsc{Resolved}$(x', x'', \clipped{z}', \clipped{z}'')$}
\label{alg:resolved}
\begin{algorithmic}[1]
\IF{$(x', x'', \clipped{z}', \clipped{z}'') \neq \text{null}$}
    \STATE $y' \gets \textsc{CEProp}(x', \clipped{z}'), \; y'' \gets \textsc{CEProp}(x'', \clipped{z}'')$
    \STATE $L_{\text{prop}} \gets \lambda \max(0, \max(\underline{\varepsilon} - y' + y'', y' - y'' - \bar{\varepsilon}))$
    \IF{$L_{\text{prop}} > 0$}
        \RETURN False
    \ENDIF
\ENDIF
\RETURN True
\end{algorithmic}
\end{algorithm}

%% file: tables/architectures.tex
\begin{table}[tb]
\centering
\begin{tabular}{@{}l@{\hspace{4pt}}l@{}}
\toprule
\multicolumn{2}{c}{\emph{Monotonicity}} \\
\midrule
\multirow{2}{*}{$h$} & $\text{relu}_{16} \to \text{relu}_{32} \to \text{relu}_{64}$ \\
                     & $\to \text{relu}_{32} \to \text{relu}_{16} \to \text{linear}_{8}$ \\
$\underline{h},\overline{h}$ & $\text{relu}_{32} \to \text{linear}_{8}$ \\
$g$ & $\text{linear}_{1}$ \\

\toprule
\multicolumn{2}{c}{\emph{Robustness}} \\
\midrule
\multirow{2}{*}{$h$} & $\text{conv}_{8} \to \text{conv}_{8} \to \text{conv}_{16} \to \text{conv}_{16} \to \text{conv}_{32}$ \\
                     & $\to \text{conv}_{32} \to \text{flatten} \to \text{relu}_{32} \to \text{linear}_{8}$ \\
$\underline{h},\overline{h}$ & $\text{linear}_{8}$ \\
$g$ & $\text{linear}_{1}$ \\

\toprule
\multicolumn{2}{c}{\emph{Fairness}} \\
\midrule
$h$ & $\text{relu}_{32} \to \text{relu}_{32} \to \text{linear}_{8}$ \\
$\underline{h},\overline{h}$ & $\text{relu}_{4} \to \text{linear}_{8}$ \\
$g$ & $\text{linear}_{1}$ \\
\bottomrule
\end{tabular}
\caption{Neural architectures.}
\label{tab:architectures}
\end{table}

%% file: tables/training_hyperparameters.tex
\begin{table}[h]
    \centering
    \renewcommand{\arraystretch}{1.}
    \begin{tabular}{ccc}
        \hline
        \emph{Parameter} & \emph{Monotonicity} & \emph{Robustness} \\
        \hline
        optimizer        & adam  & adam \\
        loss             & mse   & bce \\
        pretr. epochs    & 1000  & 20 \\
        pretr. stop patience & 10    & - \\
        epochs           & 100   & 20    \\
        batch size       & 64   & 256 \\
        \hline
    \end{tabular}
    \caption{Training hyperparameters for \emph{robustness} and \emph{monotonicity}}
    \label{tab:training_hyperparameters}
\end{table}

%% file: tables/training_hyperparameters_fairness.tex
\begin{table}[h]
    \centering
    \renewcommand{\arraystretch}{1.}
    \begin{tabular}{cccc}
        \hline
        & \multicolumn{3}{c}{\emph{Fairness}} \\
        \hline
        \emph{Parameter} & \emph{Compas} & \emph{Crime} & \emph{Law} \\
        \hline
        optimizer        & adam  & adam  & adam \\
        loss             & bce   & mse   & bce \\
        pretr. epochs    & 50    & 50  & 50 \\
        pretr. stop patience & -     & -   & - \\
        epochs           & 50  & 50  & 50 \\
        batch size       & 256   & 128   & 1024 \\
        \hline
    \end{tabular}
    \caption{Training hyperparameters for \emph{fairness}}
    \label{tab:training_hyperparameters_fairness}
\end{table}